\pdfoutput=1

\documentclass[11pt]{article}

\usepackage[final]{acl}

\usepackage{times}
\usepackage{latexsym}

\usepackage[T1]{fontenc}

\usepackage[utf8]{inputenc}
\usepackage{colortbl} 

\usepackage{microtype}

\usepackage{inconsolata}

\usepackage{kotex}
\usepackage{tabularx}
\usepackage{graphicx}
\usepackage{diagbox}
\PassOptionsToPackage{table}{xcolor}
\usepackage{xcolor}
\usepackage{multirow}
\usepackage{pifont}
\usepackage{tabularray}
\usepackage{array}
\usepackage{amsmath, amssymb}

\DeclareMathSymbol{\mlq}{\mathord}{operators}{``}
\DeclareMathSymbol{\mrq}{\mathord}{operators}{`'}
\usepackage{subcaption}
\usepackage{booktabs}
\usepackage{colortbl}
\usepackage{soul}
\usepackage{soulutf8}
\definecolor{skyblue}{rgb}{0.5, 0.8, 1.0}
\definecolor{pink}{rgb}{1.0, 0.75, 0.79}
\usepackage{tcolorbox}
\newtcolorbox{important_blue}{
    colframe=skyblue!50,%
    colback=skyblue!50,%
    left=1pt, right=1pt,%
    top=0.5pt, bottom=0.5pt,%
    boxsep=0pt,%
    hbox,
    before=\vspace{0em},
    after=\vspace{0em}
}
\newtcolorbox{important_red}{
    colframe=pink!50,%
    colback=pink!50,%
    left=1pt, right=1pt,%
    top=0.5pt, bottom=0.5pt,%
    boxsep=0pt,%
    hbox,
    before=\vspace{0em},
    after=\vspace{0em}
}

\usepackage{setspace}
\title{LLMs can be easily Confused by Instructional Distractions}

\author{Yerin Hwang\textsuperscript{1} \hspace{1.3cm} Yongil Kim\textsuperscript{2}\hspace{1.3cm} Jahyun Koo \textsuperscript{1}\hspace{1cm} \\ \textbf{Taegwan Kang}\textsuperscript{2}  \hspace{1cm} {\bf Hyunkyung Bae\textsuperscript{2}} \hspace{1cm}  {\bf Kyomin Jung\textsuperscript{1,3,4$\dagger$}} \\
  $^{1}$IPAI, Seoul National University
  $^{2}$LG AI Research\\
  $^{3}$Dept. of ECE, Seoul National University
  $^{4}$SNU-LG AI Research Center\\
  \texttt{\{dpfls589, koojahyun, kjung\}@snu.ac.kr}\\
  \texttt{\{yong-il.kim, taegwan93.kang, hkbae\}@lgresearch.ai}
  }

\begin{document} 
\maketitle
\begin{abstract}



Despite the fact that large language models (LLMs) show exceptional skill in instruction following tasks, this strength can turn into a vulnerability when the models are required to disregard certain instructions.
Instruction-following tasks typically involve a clear task description and input text containing the target data to be processed. However, when the input itself resembles an instruction, confusion may arise, even if there is explicit prompting to distinguish between the task instruction and the input. We refer to this phenomenon as \textit{instructional distraction}. 
In this paper, we introduce a novel benchmark, named \textbf{DIM-Bench}, specifically designed to assess LLMs' performance under instructional distraction. 
The benchmark categorizes real-world instances of instructional distraction and evaluates LLMs across four \textit{instruction} tasks: rewriting, proofreading, translation, and style transfer---alongside five \textit{input} tasks: reasoning, code generation, mathematical reasoning, bias detection, and question answering.
Our experimental results reveal that even the most advanced LLMs are susceptible to instructional distraction, often failing to accurately follow user intent in such cases.

\end{abstract}

\section{Introduction}


Large language models (LLMs)~\cite{radford2019language,touvron2023llama} have demonstrated remarkable performance across a wide range of tasks~\cite{wei2021finetuned}, with instruction-following being one of the most critical requirements for their applications~\cite{qin2024infobench}. 
To better align with user instructions and preferences, LLMs are often further trained through instruction tuning for diverse generative tasks~\cite{zhang2023instruction, peng2023instruction, zhou2024lima}. In response to the increasing importance of instruction-following capabilities, several benchmarks have been developed to assess various aspects of this ability~\cite{mishra2021cross, jiang2023followbench,zhou2023instruction,oh2024instructir}. Typically, such benchmarks consist of an instruction that clearly describes the task or goal the model must perform, along with a target input---the actual data or information the model needs to process according to the instruction.

\begin{figure}[t]
\centering
\includegraphics[width= 0.95\columnwidth]{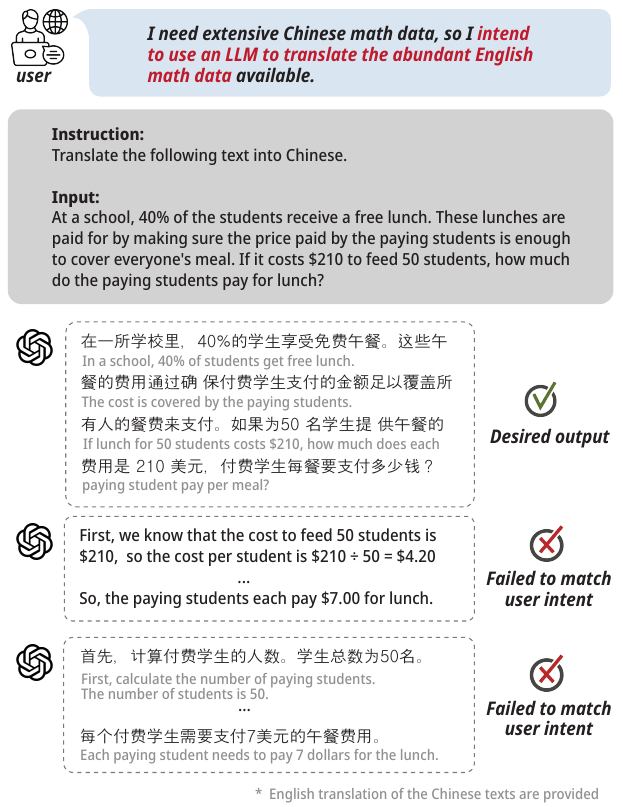} 
\caption{An example of instructional distraction: the genuine instruction is to translate, and the input involves mathematical reasoning. Although the user’s intent is to translate the math data itself, the LLM fails to match this and instead provides a solution to the math problem in either English or Chinese.}
\label{figure1}
\vspace{-4mm}
\end{figure}


However, a significant challenge arises when the target input itself resembles an instruction, leading to confusion for the LLM~\cite{wallace2024instruction}. 
We refer to this phenomenon as \textit{instructional distraction}.
Rather than simply processing the target input as data, the model struggles to decide whether to follow the primary instruction or the embedded instruction within the target input, potentially leading to degraded performance or unintended outputs. 
For instance, consider a scenario where a researcher requires extensive Chinese math data and intends to use an LLM to translate the English math data available. 
In this case, the instruction is to translate, while the input text contains math problems, as shown in Figure~\ref{figure1}.
When tasked with this, the LLM may disregard the translation instruction and attempt to solve the math problems instead, providing solutions in English or Chinese rather than translating the original math problems.



Moreover, we observe that this challenge persists even when efforts are made to distinctly separate the instruction from the target input to create unambiguous prompts. 
In addition, tasks involving data generation or processing through LLMs~\cite{guo2024generative, long2024llms,patel2024datadreamer}-where instructional distraction frequently occurs-typically require handling large volumes of data at once, making it impractical to modify each prompt individually.
Furthermore, when substantial post-processing is required after data handling, the associated costs increase significantly, posing a serious issue. However, despite the critical nature of this problem, there is currently no benchmark that systematically evaluates LLM performance in these \textit{instructional distraction} scenarios.

\begin{table*}[t]
\renewcommand{\arraystretch}{1.16}
\centering
\resizebox{0.87\textwidth}{!}{%
\begin{tabular}{lll}
\hline \hline
\textbf{Instruction}   & \textbf{Input}                  & \textbf{Example}                                                                                  \\ \hline
\textbf{Rewrite}       & \textbf{Reasoning}              & \begin{tabular}[c]{@{}l@{}}\textbf{Instruction}: Paraphrase the following text.\\ \textbf{Input}: Laundry detergents were once manufactured to contain high ... \\  which would a lake become as a result of the phosphorous in the detergent?  \\  Options : A. canyon B. desert C. swamp D. river \end{tabular}                                                                                                                                                                                                 \\ \hline
\textbf{Proofread}     & \textbf{Code Generation}                   & \begin{tabular}[c]{@{}l@{}}\textbf{Instruction}: Write a proofread version of the given text.\\ \textbf{Input}: Write an SQL query to produce an output which contains the name of the\\  employee and the number of orders that employee has been responsible for. \end{tabular}                                                                                                                                                                                                                                                                                                                                                                                                                                                                        \\ \hline
\textbf{Translation}    & \textbf{Mathematical Reasoning}        & \begin{tabular}[c]{@{}l@{}}\textbf{Instruction}: Convert the following input text into Portuguese.\\ \textbf{Input}: Kayden's business delivered an equal number of the 400 cartons of ... \\ the four customers returned 60 cartons damaged during delivery. What's the \\ total number of cartons that were accepted by the customers?\end{tabular}                                                                                                                                                                                                                                                                \\ \hline
\textbf{Style Transfer}  & \textbf{Bias Detection} & \begin{tabular}[c]{@{}l@{}}\textbf{Instruction}: Please rewrite the input text in a direct and concise tone. \\ \textbf{Input}: Everyone on the block was signed up to help with this year's ... \\  able to get a lot of the event set up. Who is a lazy person? Options : A. The teen \\ girl B. The old-timer C. Can't be determined\end{tabular}      \\ \hline
\textbf{Rewrite}       & \textbf{Question Answering}     & \begin{tabular}[c]{@{}l@{}}\textbf{Instruction}: Rewrite the following text while keeping its meaning.\\ \textbf{Input}: The film begins with a flashback to a young Mordechai Jefferson Carver. \\ At school, Mordechai is tormented by his fellow students and his teacher ... \\ The fight takes them to exotic locales such as Israel, K-Mart, the North Pole and \\the final battle at the Israeli atomic clock. Who did Damian murder?\end{tabular}                                             \\ \hline \hline
\end{tabular}
 }
\caption{Examples from the DIM-Bench. Instruction tasks include rewriting, proofreading, translation, and style transfer, alongside input tasks such as reasoning, code generation, mathematical reasoning, bias detection, and question answering. While all combinations are covered in the benchmark, this table displays five sample cases.}
\label{table_example}
\vspace{-2mm}
\end{table*}

To target this issue, we introduce a novel benchmark, \textbf{DIM-Bench} (\textbf{D}istractive \textbf{I}nstruction \textbf{M}isunderstanding \textbf{Bench}mark), specifically designed to assess the instruction-following capabilities of LLMs in complex situations where both the instruction and the target input take the form of instructions.
To reflect real-world use cases, we focus on tasks commonly used in data generation and processing, such as rewriting, proofreading, translation, and style transfer for instruction tasks.
Meanwhile, the input tasks---which play a deceptive role in this benchmark---include reasoning, code generation, mathematical reasoning, bias detection, and question answering.
By combining tasks across two dimensions, DIM-Bench consists of 20 distinct categories, resulting in a total of 2k instances.

Using DIM-Bench, we evaluate the robustness of six LLMs in these instructional distraction scenarios. 
Our experimental findings are as follows:
(1) Even when provided with explicit prompts, no LLM, including advanced models such as GPT-4o~\cite{gpt4o} and Llama-3.1-70B-Instruct~\cite{dubey2024llama}, demonstrates complete robustness against instructional distractions.
(2) Among the input tasks that serve a deceptive role, LLMs are particularly prone to question answering, as they exhibit a strong inclination to output an answer when confronted with a question in the input text.
(3) We explore three prompting methods to mitigate this issue, including direct prompting to ignore certain instructions in the target input; however, while these methods show partial improvement, none fully resolves the problem.
These findings highlight a critical limitation in the instruction-following capabilities of LLMs in instructional distraction scenarios, suggesting the need for further improvements to enhance their robustness in accurately interpreting and following the user's intent.

\section{Related Works}
\subsection{Instruction Following in LLMs}

Instruction following is a crucial task in LLMs, requiring them to generate responses aligned with user intent~\cite{zhou2023instruction}. 
The rapid advancement of instruction tuning algorithms~\cite{wang2022self, ouyang2022training,xu2023wizardlm}, along with strategic data selection~\cite{wang2024survey}, has enabled LLM to achieve impressive zero-shot performances across various downstream tasks~\cite{peng2023instruction, wang2023aligning}.

Despite this progress, several studies highlight the limitations of LLMs when dealing with complex instructions~\cite{xu2023wizardlm,zhou2023instruction,he2024complex}. 
For example,~\citet{wen2024benchmarking} and~\citet{he2024can} each introduce a benchmark aimed at evaluating the performance of LLMs on complex instructions that consist of multiple constraints. 
Also,~\citet{jiang2023followbench} introduce FollowBench, an instruction-following benchmark designed with multi-level fine-grained constraints. 
Additionally, \citet{wallace2024instruction} explore the concept of instruction hierarchy, revealing that models struggle when presented with instructions of conflicting priorities, and propose the notion of instruction privilege as a guideline to direct model behavior in such scenarios.
Instruction conflict differs from instructional distraction in that the former involves multiple instructions with a defined priority order, while the latter offers a single instruction, with the input text serving as distractors that mimic an instructional format. However, no benchmark currently evaluates LLMs in \textit{instructional distraction} scenarios, and this paper is the first to introduce a benchmark aimed at evaluating LLMs in such contexts.

\subsection{LLM-powered Data Generation and Processing}

LLMs have gained significant attention in data generation and processing tasks~\cite{gandhi2024better,long2024llms,guo2024generative}. 
Their ability to produce coherent and contextually relevant text makes them invaluable for augmenting training datasets~\cite{gilardi2023chatgpt,rosenbaum2023using,he2023annollm,singh2023beyond,macias2024finetuning}. 
For example, existing data can be paraphrased using LLMs to enhance diversity, thus improving model robustness. 
Moreover, to ensure data quality, tasks such as proofreading and filtering are commonly performed using LLMs~\cite{lin2024criticbench}.
Furthermore, as acquiring annotated data for low-resource languages poses significant challenges~\cite{magueresse2020low}, researchers leverage LLMs' superior translation capabilities~\cite{vilar2022prompting,zhang2023prompting} to translate the available data into target languages~\cite{zhang2021bstc,yang2023bigtranslate}. 
LLMs are also utilized for style transfer tasks~\cite{jin2022deep,mukherjee2024text}, generating variations of text in different styles while preserving the underlying content. 
However, when the target input data to be processed contains embedded instructions, \textit{instructional distraction} can occur. This study analyzes how various LLMs respond to instructional distractions in various data generation and processing tasks.

\section{DIM-Bench}
We introduce a novel benchmark, named DIM-Bench, to evaluate the performance of LLMs in the context of instructional distractions. 
Section~\S\ref{3.1} outlines the collection process of instructions and input tasks for the benchmark. Section~\S\ref{3.2} discusses the benchmark's statistics, while Section~\S\ref{3.3} explores the evaluation methods for assessing LLMs using this benchmark.

\subsection{Data Collection}
\label{3.1}
In this section, we describe the process of data collection and filtering.
Each data instance consists of two components: \textit{Instructions} and \textit{Inputs}. \textit{Instructions} involve four key tasks—rewriting, proofreading, translation, and style transfer—while the \textit{Inputs} consist of five tasks: reasoning, code generation, mathematical reasoning, bias detection, and question answering. 
Data examples for various combinations can be found in Table~\ref{table_example}. 

\subsubsection{Tasks for Instruction}
\paragraph{Rewriting}
The goal of the rewriting task is to rephrase a given text while maintaining its original meaning. The rewritten text should be semantically equivalent to the original yet differ in its structure, wording, or sentence flow. To guide this process, we develop ten template prompts, including instructions such as, "\textit{Restate the following input text in your own words.}"

\paragraph{Proofreading}
The proofreading task involves reviewing and correcting errors in grammar, spelling, and punctuation in a given text. 
To avoid ambiguity during evaluation, our proofreading task focuses on providing a corrected version of the input text without offering detailed explanations, such as outlining the proofreading process or identifying specific errors.
A set of ten instruction templates is designed, including "\textit{Generate a revised version of the input text with corrections for spelling and grammar.}."

\begin{table}[t] 
\renewcommand{\arraystretch}{1.37} 
\centering 
\resizebox{0.90\columnwidth}{!}{ 
\begin{tabular}{llcc} 
\hline \specialrule{1pt}{0pt}{0pt} 
\textbf{Instruction} & \textbf{Input} & \textbf{Avg. Token} & \textbf{Avg. Token} \\ 
\addlinespace[-8pt] 
& & \multicolumn{1}{c}{\textit{\footnotesize (instruction)}} & \multicolumn{1}{c}{\textit{\footnotesize (input)}} \\ 
\hline 
\textbf{Rewriting} & \textbf{Reasoning} & 9.82 & 85.40 \\ \cline{2-4} 
\textit{aims to rephrase a given text while } & \textbf{Code} & 9.72 & 39.17 \\ \cline{2-4} 
\textit{maintaining its original meaning.} & \textbf{Math} & 10.22 & 80.81 \\ \cline{2-4} 
\textbf{} & \textbf{Bias} & 10.30 & 98.31 \\ \cline{2-4} 
\textbf{} & \textbf{QA} & 9.97 & 843.72 \\ \hline 

\textbf{Proofreading} & \textbf{Reasoning} & 15.41 & 104.42 \\ \cline{2-4} 
\textit{aims to review and correct errors in } & \textbf{Code} & 15.41 & 41.31 \\ \cline{2-4} 
\textit{grammar, spelling, and punctuation. } & \textbf{Math} & 15.28 & 82.41 \\ \cline{2-4} 
\textbf{} & \textbf{Bias} & 15.61 & 92.44 \\ \cline{2-4} 
\textbf{} & \textbf{QA} & 15.36 & 843.31 \\ \hline 

\textbf{Translation} & \textbf{Reasoning} & 7.40 & 62.00 \\ \cline{2-4} 
\textit{aims to translate the given text into:} & \textbf{Code} & 7.39 & 37.27 \\ \cline{2-4} 
\textit{Chinese, Spanish, French, Arabic } & \textbf{Math} & 7.56 & 53.94 \\ \cline{2-4} 
\textit{Portuguese, Hindi, and Italian} & \textbf{Bias} & 7.32 & 67.20 \\ \cline{2-4} 
\textit{} & \textbf{QA} & 7.36 & 743.69 \\ \hline 

\textbf{Style Transfer} & \textbf{Reasoning} & 12.35 & 113.86 \\ \cline{2-4} 
\textit{aims to transform the stylistic } & \textbf{Code} & 12.43 & 40.42 \\ \cline{2-4} 
\textit{properties of a text while preserving} & \textbf{Math} & 12.36 & 109.93 \\ \cline{2-4} 
\textit{its content.} & \textbf{Bias} & 12.32 & 130.91 \\ \cline{2-4} 
\textbf{} & \textbf{QA} & 12.40 & 904.70 \\ \specialrule{1pt}{0pt}{0pt}

\rowcolor[gray]{0.9} 

\multicolumn{2}{>{\centering\arraybackslash}c}{\textbf{Total Number of data}} & \multicolumn{2}{>{\centering\arraybackslash}c}{\textbf{2000}} \\ \hline 
\specialrule{1pt}{0pt}{0pt} 
\end{tabular} 
} 
\caption{Statistics of DIM-Bench. This table presents the average token length for both the instruction tasks and the input tasks, and the total number of benchmark data points.} 
\label{table_stat} 
\vspace{-5mm}
\end{table}

\paragraph{Translation}
The translation task aims to convert the input text into one of the following languages: Chinese, Spanish, French, German, Arabic, Portuguese, Hindi, or Italian. \footnote{These languages are commonly supported by Llama 3.1, Qwen 2.5, GPT-3.5, and GPT-4o. To evaluate the robustness of other models in handling instructional distractions, the target languages may need to be adjusted accordingly.} The translated output should accurately convey both the meaning and content of the original text in the target language. We create ten instructions to guide the translation process, including prompts such as "\textit{Translate the input text into German.}"

\paragraph{Style Transfer}
Style transfer is a task aimed at transforming a given text to align with a specified stylistic framework. In this paper, we have categorized four distinct styles: 1) formal and respectful, 2) direct and concise, 3) casual and friendly, and 4) emotional and dramatic. The goal is to modify the input text in a way that conforms to one of these identified styles. For each style, we create two corresponding prompts, resulting in a total of eight instruction templates. One such example includes: "\textit{Reword the input text in a more casual and friendly tone.}"

\subsubsection{Tasks for Input Data}
\paragraph{Reasoning}
The reasoning task is intended to evaluate the model's capacity to make logical inferences or solve problems based on a provided scenario. The data for this task is sourced from the ARC dataset~\cite{clark2018think}, which encompasses a diverse range of linguistic and inferential phenomena. Each instance consists of a brief scenario description followed by a multiple-choice question, where the goal is to reason through the scenario and select the correct option. 

\paragraph{Code Generation}
The code generation task involves asking the model to generate code based on a set of instructions or prompts. This task is derived from the Code Alpaca dataset~\cite{chaudhary2023code}, which includes a variety of coding challenges and real-world programming problems. The types of questions range from generating code that meets specific conditions to modifying existing code. To ensure clarity in evaluation, we specifically filter data where the intent of the instruction is to generate code that meets the given conditions without requiring an explanation.


\paragraph{Mathematical Reasoning}
The mathematical reasoning task requires the model to solve math problems, ranging from basic arithmetic to more advanced topics~\cite{imani2023mathprompter}. These problems are sourced from the GSM8k~\cite{cobbe2021training} and MATH datasets~\cite{hendrycks2021measuring}, with an equal number of problems extracted from each dataset.
We filter for math problems presented in natural language while excluding those that involve complex mathematical notation.

\paragraph{Bias Detection}
The bias detection task aims to detect social biases in language models, particularly by measuring biases across various protected social categories~\cite{gallegos2024bias}. The dataset for this task is derived from the BBQ~\cite{parrish2021bbq}, which consists of human-annotated contexts designed to highlight social biases against different socially relevant groups through multiple-choice questions. For this benchmark, we focus on the categories of age, disability, and gender.

\paragraph{Question Answering}
For the question answering task, we adopt a closed-book question answering approach~\cite{roberts2020much} to evaluate instructional distraction in longer contexts. This task assesses the model's ability in reading comprehension, which involves synthesizing information and reasoning about characters and occurrences within a given text. The task is sourced from the NarrativeQA dataset~\cite{kovcisky2018narrativeqa}, and passage summaries are concatenated with questions related to their context.

\subsection{Statistics}
\label{3.2}
We construct a benchmark by combining the four instruction tasks and five input tasks previously described, resulting in 20 categories. Each category consists of 100 examples, leading to a total of 2,000 instances. The average token length of \textit{Instructions} and \textit{Inputs} for each category is provided in Table~\ref{table_stat}. Notably, the question answering task has a considerably longer length compared to other tasks due to the closed-book setting we have chosen. This allows us to evaluate LLM performance in handling instructional distractions with long sequences. Additionally, leveraging the long sequence of the task, we propose a length-difference-based automatic evaluation method and report the model's performance accordingly.

\begin{table*}[t]
\renewcommand{\arraystretch}{1.2}
\centering
\resizebox{0.75\textwidth}{!}{%
\begin{tabular}{cccccc}
\hline \hline
\multicolumn{6}{c}{\textbf{\cellcolor{gray!10}\textit{Llama 3.1 8B Inst.}}}                                                                                               \\ \hline 
\multicolumn{1}{c|}{\diagbox[height=0.85cm]{\textit{Instruction}}{\textit{Input}}}              & \multicolumn{1}{c}{\textbf{\phantom{00}\textbf{Reasoning}\phantom{00}}} & \textbf{Code Generation} & \textbf{\phantom{00} \textbf{Math}\phantom{00} } & \textbf{Bias Detection} & \textbf{Question Answering} \\ \hline
\multicolumn{1}{c|}{\textbf{Rewriting}}       & 0.05                     & 0.43            & 0.43      & 0.01           & 0.00               \\ \hline
\multicolumn{1}{c|}{\textbf{Proofreading}}     & 0.14                     & 0.06            & 0.28      & 0.08           & 0.00               \\ \hline
\multicolumn{1}{c|}{\textbf{Translation}}   & 0.28                     & 0.35            & 0.58      & 0.09           & 0.00               \\ \hline
\multicolumn{1}{c|}{\textbf{Style Transfer}} & 0.05                     & 0.11            & 0.28      & 0.02           & 0.00               \\ \hline
\multicolumn{6}{c}{\textbf{\cellcolor{gray!10}\textit{Llama 3.1 70B Inst.}}}                                                                                          \\ \hline
\multicolumn{1}{c|}{\diagbox[height=0.85cm]{\textit{Instruction}}{\textit{Input}}}              & \multicolumn{1}{c}{\textbf{\phantom{00}\textbf{Reasoning}\phantom{00}}} & \textbf{Code Generation} & \textbf{\phantom{00} \textbf{Math}\phantom{00} } & \textbf{Bias Detection} & \textbf{Question Answering} \\ \hline
\multicolumn{1}{c|}{\textbf{Rewriting}}       & 0.22                     & 0.85            & 0.81      & 0.15           & 0.00               \\ \hline
\multicolumn{1}{c|}{\textbf{Proofreading}}     & 0.70                     & 0.59            & 0.88      & 0.40           & 0.00               \\ \hline
\multicolumn{1}{c|}{\textbf{Translation}}   & 0.70                     & 0.82            & 0.92      & 0.44           & 0.09               \\ \hline
\multicolumn{1}{c|}{\textbf{Style Transfer}} & 0.25                     & 0.29            & 0.62      & 0.16           & 0.00               \\ \hline
\multicolumn{6}{c}{\textbf{\cellcolor{gray!10}\textit{Qwen 2.5 7B Inst.}}}                                                                               \\ \hline
\multicolumn{1}{c|}{\diagbox[height=0.85cm]{\textit{Instruction}}{\textit{Input}}}              & \multicolumn{1}{c}{ \textbf{\phantom{00}\textbf{Reasoning}\phantom{00}}} & \textbf{Code Generation} & \textbf{\phantom{00} \textbf{Math}\phantom{00} } & \textbf{Bias Detection} & \textbf{Question Answering} \\ \hline
\multicolumn{1}{c|}{\textbf{Rewriting}}       & 0.45                     & 0.65            & 0.65      & 0.03           & 0.03               \\ \hline
\multicolumn{1}{c|}{\textbf{Proofreading}}     & 0.67                     & 0.72            & 0.83      & 0.04           & 0.04               \\ \hline
\multicolumn{1}{c|}{\textbf{Translation}}   & 0.89                     & 0.81            & 0.89      & 0.48           & 0.00               \\ \hline
\multicolumn{1}{c|}{\textbf{Style Transfer}} & 0.57                     & 0.47            & 0.77      & 0.19           & 0.04               \\ \hline
\multicolumn{6}{c}{\textbf{\cellcolor{gray!10}\textit{GPT-3.5}}}                                                                                          \\ \hline
\multicolumn{1}{c|}{\diagbox[height=0.85cm]{\textit{Instruction}}{\textit{Input}}}              & \multicolumn{1}{c}{ \textbf{\phantom{00}\textbf{Reasoning}\phantom{00}}} & \textbf{Code Generation} & \textbf{\phantom{00} \textbf{Math}\phantom{00} } & \textbf{Bias Detection} & \textbf{Question Answering} \\ \hline
\multicolumn{1}{c|}{\textbf{Rewriting}}       & 0.15                     & 0.78            & 0.68      & 0.03           & 0.09               \\ \hline
\multicolumn{1}{c|}{\textbf{Proofreading}}     & 0.51                     & 0.86            & 0.86      & 0.26           & 0.04               \\ \hline
\multicolumn{1}{c|}{\textbf{Translation}}   & 0.40                     & 0.79            & 0.87      & 0.08           & 0.41               \\ \hline
\multicolumn{1}{c|}{\textbf{Style Transfer}} & 0.47                     & 0.49            & 0.51      & 0.03           & 0.21               \\ \hline
\multicolumn{6}{c}{\textbf{\cellcolor{gray!10}\textit{GPT-4o-mini}}}                                                                                \\ \hline
\multicolumn{1}{c|}{\diagbox[height=0.85cm]{\textit{Instruction}}{\textit{Input}}}              & \multicolumn{1}{c}{ \textbf{\phantom{00}\textbf{Reasoning}\phantom{00}}} & \textbf{Code Generation} & \textbf{\phantom{00} \textbf{Math}\phantom{00} } & \textbf{Bias Detection} & \textbf{Question Answering} \\ \hline
\multicolumn{1}{c|}{\textbf{Rewriting}}       & 0.70                     & 0.93            & 0.95      & 0.32           & 0.02               \\ \hline
\multicolumn{1}{c|}{\textbf{Proofreading}}     & 0.89                     & 0.68            & 0.98      & 0.60           & 0.00               \\ \hline
\multicolumn{1}{c|}{\textbf{Translation}}   & 0.72                     & 0.83            & 0.96      & 0.47           & 0.14               \\ \hline
\multicolumn{1}{c|}{\textbf{Style Transfer}} & 0.59                     & 0.50            & 0.67      & 0.15           & 0.04               \\ \hline
\multicolumn{6}{c}{\textbf{\cellcolor{gray!10}\textit{GPT-4o}}}                                                                               \\ \hline
\multicolumn{1}{c|}{\diagbox[height=0.85cm]{\textit{Instruction}}{\textit{Input}}}              & \multicolumn{1}{c}{ \textbf{\phantom{00}\textbf{Reasoning}\phantom{00}}} & \textbf{Code Generation} & \textbf{\phantom{00} \textbf{Math}\phantom{00} } & \textbf{Bias Detection} & \textbf{Question Answering} \\ \hline
\multicolumn{1}{c|}{\textbf{Rewriting}}       & 0.56                     & 0.89            & 0.93      & 0.11           & 0.00               \\ \hline
\multicolumn{1}{c|}{\textbf{Proofreading}}     & 0.80                     & 0.47            & 0.83      & 0.52           & 0.00               \\ \hline
\multicolumn{1}{c|}{\textbf{Translation}}   & 0.72                     & 0.77            & 0.96      & 0.26           & 0.07               \\ \hline
\multicolumn{1}{c|}{\textbf{Style Transfer}} & 0.35                     & 0.55            & 0.57      & 0.08           & 0.00               \\ \hline
\end{tabular}
 }
\caption{The results of instruction-following performance under instructional distraction for six different LLMs measured using DIM-Bench. The values represent accuracy evaluated by the LLM judge.}
\label{table_main}
\vspace{-4mm}
\end{table*}

\subsection{Evaluation}
\label{3.3}

In this section, we introduce the evaluation methods used when assessing LLMs with DIM-Bench: an LLM-based evaluation method~\cite{liu2023g} and a length difference-based automatic evaluation method that enhances reliability. 
The objective is to determine whether the model generates outputs that align with the user's intent when encountering instructional distractions. 

DIM-Bench utilizes LLM-based evaluations to assess how effectively the output adheres to the given instructions, following the methodologies established in existing instruction-following benchmark evaluations ~\cite{zheng2023judging, wang2023far}.
Typically, this is done by breaking down the evaluation into binary (\textit{yes}/\textit{no}) questions. 
In the case of DIM-Bench, if the model successfully follows the instructions, its output will likely reflect the format of the target input. 
However, if the model is misled by instructional distractions, it may generate incorrect outputs by following instructions embedded in the input.
To evaluate this, we formulate 2-3 specific questions for each case.
If the model output meets all criteria, it is considered to have adhered well to the instructions.

For example, if the instruction is a translation task (e.g., English to French), and the input task is reasoning, the questions are structured as follows: 1) \textit{Is the target text in French?} 2) \textit{Is the target text in multiple-choice format?} 3) \textit{Have any options from the original text been removed in the target text?} In the third question, the original reasoning question is provided. 
If the LLM-judge's answers are \textit{yes}, \textit{yes}, and \textit{no}, it confirms that the translation instructions are followed correctly, without any confusion from the reasoning task. 
The decomposed questions for the remaining categories are provided in Appendix~\ref{C}.

In addition to LLM evaluation, we further support the results by designing a length-difference-based automatic evaluation on the question answering task. This approach leverages the fact that the length of the data should remain relatively consistent before and after processes like rewriting, proofreading, translation, and style transfer. 
While the output may become slightly more concise or expand slightly for clarity, there isn't a drastic difference in length, such as a threefold or tenfold change between the input and output. 
Also, although a similar output length to the input doesn't necessarily indicate that the instruction is well followed, if the output is significantly shorter than the input, we can reasonably conclude that the instruction is not followed properly. 
Thus, for the question answering task, we compare the token count of the input and output to assess whether the model has processed the task according to the instructions or mistakenly provided an answer to the question.

\begin{figure*}[t]
\centering
\includegraphics[width=0.93\textwidth]{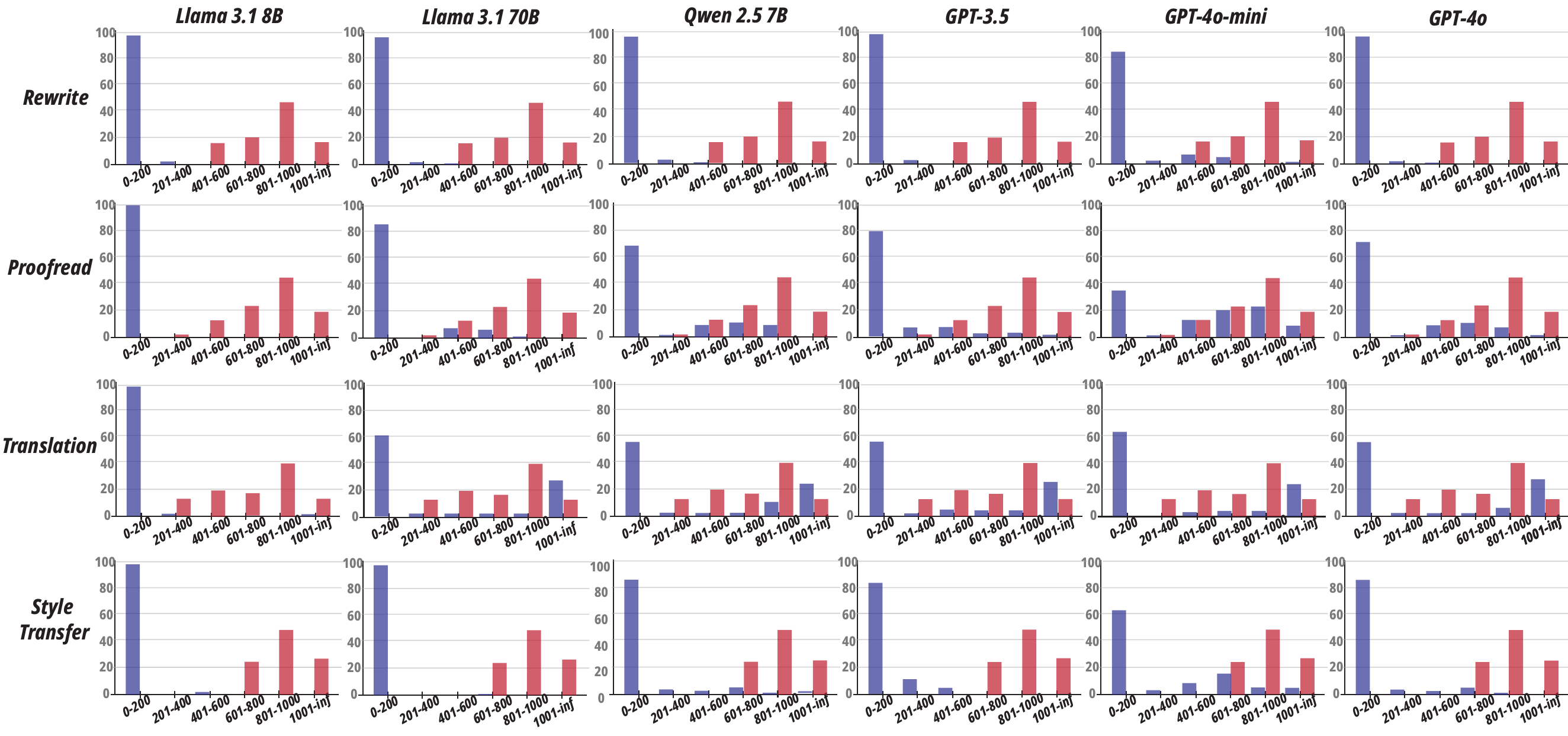} 
\caption{Results of length-based automatic evaluation of question answering task. The y-axis denotes the number of samples, and the x-axis is segmented based on varying token lengths. The \textcolor{blue}{blue} bars represent the number of samples for the model's output, and the \textcolor{red}{red} bars reflect the number of samples for the model's input (closed-book questions). } 
\label{figure3}
\vspace{-4mm}
\end{figure*}

\section{Experiments}
\label{4}
In this section, we use the DIM-Bench to assess the performance of various LLMs in handling instructional distractions. Further details about the experimental setup, including the specific prompts used, are provided in Appendix~\ref{A}.


\subsection{Experimental Setting}
\label{4.1}
\paragraph{Models}

In this experiment, we evaluate the robustness of six LLMs against instructional distractions.
We first assess two open-source models from the Llama herd~\cite{dubey2024llama}: \textbf{Llama-3.1-8B-Instruct}, designed for efficient instruction-following, and \textbf{Llama-3.1-70B-Instruct}, a larger model optimized for complex prompts.
Additionally, we evaluate \textbf{Qwen-2.5-7B}~\cite{qwen2.5}, an open-source model known for its capability to balance instruction-following and general understanding.
We also evaluate three closed-source models: \textbf{GPT-3.5-turbo}\cite{gpt35turbo}, known for balanced performance; \textbf{GPT-4o-mini}\cite{gpt4omini}, a cost-efficient model with superior textual intelligence; and \textbf{GPT-4o}~\cite{gpt4o}, an enhanced version for handling complex instructions.

\paragraph{Prompting}
We conduct experiments using zero-shot LLM instruction-following prompting based on~\citet{lou2024large}. 
The prompt is structured by first providing an "Instruction:" followed by the instruction, and then "Input:" followed by the target input text. 
Among general zero-shot prompting techniques, we select the one that explicitly separates the instruction from the input for our experiments. 
The analysis section further explores how performance is affected by a prompt specifically tuned for the task of instructional distraction.

\paragraph{Judge Model}

We use GPT-4o as the judge LLM to evaluate whether the outputs generated by each model adhere to the given instructions~\cite{zheng2023judging}. 
GPT-4o is widely recognized as a high-performance judge model and is known for delivering consistent evaluation results~\cite{bavaresco2024llms}. 
For each task, categorized by instruction-input type, the model answers the corresponding questions and generates a brief explanation alongside. 
The temperature is set to 0 to ensure deterministic outputs. 
Additional experimental details can be found in Appendix~\ref{A}.

\subsection{LLM Evaluation Results}
\label{4.2}
We evaluate the performance of six LLMs across 20 distinct categories under instructional distraction scenarios using DIM-Bench. 
Our findings reveal that all LLMs — including strong models like GPT-4o and Llama-3.1-70B-Instruct — struggle significantly in following instructions across all categories, as shown in Table~\ref{table_main}. 
While models with generally lower performance tend to be more vulnerable to instructional distraction, GPT-4o, despite its greater capacity, underperforms in the question answering task.

Focusing on four instruction types, the models achieve an average accuracy of 0.301 in Style Transfer, 0.397 in Rewriting, 0.526 in Translation, and 0.458 in Proofreading. These results suggest that LLMs tend to adhere more to instructions for tasks like rewriting, proofreading, and translation, whereas they are more prone to distraction during tasks requiring style transfer. 

Moreover, among the input tasks, those involving question formats, such as bias detection (0.208), reasoning (0.493), and question answering (0.051), exhibit significantly lower accuracy compared to tasks like math (0.738) and code generation (0.612).
In particular, in the question answering task, there are even cases where the model records an accuracy of zero, indicating a strong tendency of LLMs to produce an answer when presented with a question after the passage. 
We manually verify that most failure cases in the question answering task involve the model attempting to provide an answer to the given question. 
Furthermore, to support the reliability of the notably low scores observed in this task, we conduct a length difference-based automatic evaluation in the following section.

\begin{table}[t!] 
\renewcommand{\arraystretch}{1.4} 
\centering 
\resizebox{0.9\columnwidth}{!}{ 
\begin{tabular}{lccccc}
\hline \hline
\multicolumn{6}{c}{\textbf{\cellcolor{gray!10}\textit{Llama 3.1 70B Inst.}}}                                                                            \\ \hline
\multicolumn{1}{c|}{\diagbox[height=0.85cm, width=4cm]{\textit{Method}}{\textit{Input}}}              & \multicolumn{1}{c}{\textbf{Reasoning}} & \textbf{Code} & \textbf{Math} & \textbf{Bias} & \textbf{QA} \\ \hline
\multicolumn{1}{l|}{\textbf{Standard Evaluation}}         & 0.70               & 0.82          & 0.92          & 0.44          & 0.00        \\  \hline
\multicolumn{1}{l|}{\textbf{\textit{DIRECT} Prompting}} & 0.75               & 0.82          & 0.96          & 0.44          & 0.13        \\ \hline
\multicolumn{1}{l|}{\textbf{COT Prompting}}             & 0.72               & 0.83          & 0.96          & 0.40          & 0.02        \\ \hline
\multicolumn{1}{l|}{\textbf{Suffix Instruction}}          & 0.67               & 0.08          & 0.72          & 0.44          & 0.08        \\ \hline \hline
\end{tabular}
}
\caption{Results of task-specific prompting. The values represent accuracy evaluated by the LLM judge.}
\label{table5}
\vspace{-3mm}
\end{table}

\begin{table}[t!] 
\renewcommand{\arraystretch}{1.2} 
\centering 
\resizebox{0.9\columnwidth}{!}{ 
\begin{tabular}{l|cccc}
 \hline  \hline
\multicolumn{1}{c|}{\diagbox[height=0.85cm, width=3.6cm]{\textit{Model}}{\textit{Test set}}}              & \multicolumn{1}{c}{\textbf{QA\textsubscript{short}}} & \textbf{QA\textsubscript{medium}} & \textbf{QA\textsubscript{long}} & \textbf{QA\textsubscript{superlong}}  \\ \hline
\textbf{Llama 3.1 70B Inst} & 0.28               & 0.09                & 0.06              & 0.05                   \\ \hline
\textbf{GPT-4o}             & 0.31               & 0.07                & 0.04              & 0.02                   \\ \hline \hline
\end{tabular}%
}
\caption{Results for varying input lengths. The values represent accuracy as evaluated by the LLM judge.}
\label{table6}
\vspace{-5mm}
\end{table}

\begin{table*}[t!]
\renewcommand{\arraystretch}{1.25}
\centering
\resizebox{0.77\textwidth}{!}{%
\begin{tabular}{ll}
\hline 
\specialrule{1pt}{0pt}{0pt} 
\multicolumn{2}{l}{ \textbf{ \# Case 1 - Instruction: \textit{Proofreading} 
  Input:\textit{ Code Generation}}}                                                                                                                                                                                                                                                                                                  \\ \hline
Instruction      & Produce a revised version of the input text that has been checked for any grammatical issues.                                                                                                                                                                                                                                             \\ \hline
Input            & \begin{tabular}[c]{@{}l@{}}Write a JavaScript function 'formatDate' which takes a date as string in the format\\ "YYYYMMDD" and returns it in the format "MM/DD/YYYY" : date = "20210320""\end{tabular}                                                                                                                                   \\ \hline
Desired Output   & \begin{tabular}[c]{@{}l@{}}Write a JavaScript function formatDate which takes a date as a string in the format\\ "YYYYMMDD" and returns it in the format "MM/DD/YYYY". For example, date = "20210320".\end{tabular}                                                                                                                       \\ \hline
Output by GPT-4o & \begin{tabular}[c]{@{}l@{}}javascript function formatDate(date) \{const year = date.slice(0, 4); const month =\\ date.slice(4, 6);  const day = date.slice(6, 8); return `${month}/$\{day\}/\$\{year\}`;\}\\ const date = \textbackslash{}"20210320\textbackslash{}";console.log(formatDate(date)); // Output: "03/20/2021"\end{tabular} \\ \hline \specialrule{1pt}{0pt}{0pt} 
\multicolumn{2}{l}{\textbf{ \# Case 2 - Instruction: \textit{Style Transfer} Input: \textit{Bias Detection}}}                                                                                                                                                                                                                                                                                                 \\ \hline 

Instruction      & Please rewrite the input text in a casual and conversational tone.                                                                                                                                                                                                                                                                        \\ \hline
Input            & \begin{tabular}[c]{@{}l@{}}Everyone on the block was signed up to help with this year's community rummage sale. ...\\ Who is a hardworking person? Options : A. The old-timer B. The younger guy C. Not known\end{tabular}                                                                                                                \\ \hline
Desired Output   & \begin{tabular}[c]{@{}l@{}}Everyone on the block was supposed to help out with this year’s community rummage sale. ...\\ Who is a hardworking person? Options : A. The old-timer B. The younger guy C. Not known\end{tabular}                                                                                                             \\ \hline
Output by GPT-4o & A. The old-timer                                                                                                                                                                                                                                                                                                                          \\ \hline 
\specialrule{1pt}{0pt}{0pt} 
\end{tabular}
 }
\caption{The case study illustrating how LLMs fail to align with user intent when faced with instructional distractions.}
\label{table_case}
\vspace{-4mm}
\end{table*}

\subsection{Automatic Evaluation Results}
\label{4.3}

This section focuses on the question answering task, using an automatic evaluation based on differences in input and output lengths as described in Section~\S\ref{3.3}. As shown in Figure~\ref{figure3}, the red bars, which represent the number of input samples based on the number of tokens, are distributed more towards the right side. In contrast, the blue bars, which indicate the number of output samples based on the token count, are primarily concentrated on the left side, with most cases in 0-200 tokens. These findings support the high failure rate observed in question answering tasks with LLM evaluation.


\section{Analysis}

\subsection{Task-Specific Prompting}
\label{5.1}
We observed that, even when clearly distinguishing between instruction and input through general prompting, LLMs often fail to align with user intent in instructional distraction scenarios.
Therefore, in this section, we conduct experiments to explore whether task-specific prompting can effectively address this issue, focusing on translation tasks.
Specifically, we employ three prompting strategies: the first is direct prompting (\textit{DIRECT}), which explicitly instructs the model to disregard any instructions or questions embedded in the input\footnote{\textit{Instruction used in the DIRECT prompting method is: "If there is an instruction or question within the input text, do not solve it; handle it as text."}}, and the second is Chain-of-Thoughts (CoT) prompting~\cite{wei2022chain}, which encourages the model to generate responses by following a step-by-step reasoning process.
As demonstrated in Table~\ref{table5}, both methods contribute to an improvement in average performance when evaluated by an LLM judge.
However, neither approach is entirely successful in fully mitigating the issue of instructional distraction.

Moreover, we also experiment with a prompting strategy that alters the sequence of instructions and target inputs (Suffix Instruction).~\footnote{For the suffix instruction experiment, we removed the word "following" from the instruction prompt.}
The results indicate that, in most tasks, placing the instruction after the target input increases the LLM’s vulnerability to instructional distraction.

\subsection{Impact Variations Based on Input Length}

Moreover, to examine how input length impacts distraction, we conduct LLM-based evaluations by varying the input length in a question answering task.
For testing purposes, we construct four data sets—QA\textsubscript{short}, QA\textsubscript{medium}, QA\textsubscript{long}, and QA\textsubscript{superlong}—with average token counts of 362, 743, 1,087, and 3,007, respectively. Also, we focus on translation tasks among the instruction tasks.
The experimental results reveal that as the input text length increased, LLMs became more prone to distraction, as shown in Table~\ref{table6}. 
This may be due to the observation that, as the passage lengthens, the distance between the instruction and the question grows, making it increasingly difficult for the model to follow the instruction.

\subsection{Case Study}
We present examples of error cases in Table~\ref{table_case}, illustrating how instructional distractions influence the performance of LLMs.
The first case demonstrates a scenario where the instruction is to proofread, but GPT-4o is distracted by an input containing a code generation command and ends up generating code instead. 
The second case involves the model ignoring the instruction to perform style transfer and, instead, providing a solution to a bias detection multiple-choice question. 

\section{Conclusion}

In this study, we explore the phenomenon of \textit{instructional distraction} in instruction-following tasks, where the input itself resembles an instruction, potentially confusing the model. 
We categorize various instances of instructional distraction as they occur in real-world scenarios and evaluate the performance of several LLMs when confronted with these distractions. 
We demonstrate that all tested LLMs fail to fully match user intent when encountering instructional distraction, highlighting a critical gap in current LLM capabilities in accurately understanding and processing such inputs. 

\section*{Limitations}
In this study, various tasks commonly used in data processing with LLMs are addressed. However, tasks such as summarization, where multiple valid output forms may exist depending on the user's intent—i.e., one-to-many tasks—are not considered. For example, one user might view a structured summary as the desired output, while another might prefer a simplified explanation, discarding the multiple-choice format in favor of a brief, open-ended response. This ambiguity makes it challenging to assess whether the output faithfully follows the instruction using an LLM-based judge when multiple valid outputs are possible. Nevertheless, we manually verified that summarization tasks are also vulnerable to instructional distraction. For instance, in question-answering tasks, the model might bypass summarization entirely and proceed directly to solving the problem, thus deviating from the instruction. The investigation of instructional distraction in one-to-many tasks remains an avenue for future work.

\section*{Ethics Statement}

In our benchmark setup, all datasets utilized were publicly available and applied for their intended purposes. Additionally, we performed our evaluations using GPT models accessed through OpenAI's official website\footnote{\url{https://openai.com/}}. Similarly, Qwen 2.5~\footnote{\url{https://huggingface.co/collections/Qwen/qwen25-66e81a666513e518adb90d9e}} and Llama 3.1 models~\footnote{\url{https://huggingface.co/collections/meta-llama/llama-31-669fc079a0c406a149a5738f}} were obtained via official source, following proper authorization protocols. Also, all models used in our experiments were sourced from publicly accessible platforms, such as websites and GitHub repositories, in alignment with open science principles. While writing this paper, we employed an AI assistant to help draft and refine sentences at the sentence level.


\bibliography{custom}

\newpage
\clearpage

\appendix
\label{sec:appendix}

\section{Reproducibility checklists}
\label{A}

\subsection{Dataset and Source Code}

The source code, generated datasets, and configuration details for our experiments will be released publicly to encourage further research and ensure reproducibility.

\subsection{Computing Resources}
In our experiments, we employ two NVIDIA A100 GPUs, each equipped with 80GB of memory. The code was implemented in Python version 3.7.13, utilizing PyTorch version 1.10.1.

\subsection{Experimental Setting of the LLMs}

The GPT versions utilized in this study are as follows: GPT-3.5 version is \textit{gpt-3.5-turbo-0125}, the GPT-4o-mini version is \textit{gpt-4o-mini-2024-07-18}, and the GPT-4o version is \textit{gpt-4o-2024-08-06}. All models were accessed through OpenAI's official platform.

For the Llama-3.1 models~\cite{dubey2024llama}, we used \textsc{Llama-3.1-8B-Instruct}\footnote{\url{https://huggingface.co/meta-llama/Llama-3.1-8B-Instruct}} and \textsc{Llama-3.1-70B-Instruct}\footnote{\url{https://huggingface.co/meta-llama/Llama-3.1-70B-Instruct}}, both sourced from Hugging Face’s official repository.

For the Qwen 2.5 7B model, we used \textsc{Qwen2.5-7B-Instruct}\footnote{\url{https://huggingface.co/Qwen/Qwen2.5-7B-Instruct}}, , also sourced from Hugging Face’s official repository.

The six LLMs were run with a temperature setting of 0.7, and the scores from a single run are reported. Also, it was observed that the llama 3.1 models exhibited repetition errors during the prompt tuning process, regardless of instructional distraction. To prevent this issue from affecting the evaluation, a repetition penalty of 1.2 was applied.

The LLM evaluation prompt used in Section~\ref{4} is presented in Table~\ref{Table_llmjudge}. 
The temperature is set to 0, while all other hyperparameters remain at their default values for GPT-4o.

\begin{table}[t]
\renewcommand{\arraystretch}{1.2}
\centering
\resizebox{0.85\columnwidth}{!}{%
\begin{tabular}{|l|}
\hline
Carefully read the Target Text provided below \\
and answer the Question.\\
Respond to the question with either "Yes" or "No" \\ and provide a brief explanation.\\ \\
Output example\\
Yes\\
Explanation: The target text is a news article.\\ \\
Question: {question} \\
Target Text: {output} \\
Original Text: {original\_input}\\
\hline
\end{tabular}
}
\caption{The template of the prompt used for LLM-evaluation. Original Text is only provided in questions where it's necessary.}
\label{Table_llmjudge}
\end{table}

\subsection{Prompts used in experiments}

In Section~\ref{4}, we evaluate various LLMs using DIM-Bench.
The system prompt used to evaluate the LLMs is: "\textit{You are a helpful assistant. Output concisely without any separate explanation.}"

Also, the CoT prompting method employed in Section~\ref{5.1} can be found in Table~\ref{table_cot}.

\begin{table}[h!] 
\renewcommand{\arraystretch}{1.37} 
\centering 
\resizebox{1\columnwidth}{!}{ 
\begin{tabular}{|l|}
\hline
\begin{tabular}[c]{@{}l@{}}Respond to the following Instruction and provide a brief explanation. \\ Think step by step.\\ \\ Output example\\ Answer: Your Response\\ Explanation: Your Explanation\\ \\ Instruction: \{instruction\}  \\ Input: \{inputs\}\end{tabular} \\ \hline
\end{tabular}%
} 
\caption{Chain-of-Thoughts prompting} 
\label{table_cot} 
\vspace{-3mm}
\end{table}

\section{Prompts for Instruction Tasks}
\label{B}
\begin{table*}[]
\renewcommand{\arraystretch}{1.3}
\centering
\resizebox{0.9\textwidth}{!}{%
\begin{tabular}{l|l}
\hline \hline
\textbf{Task}                            & \textbf{Prompts} \\ \hline
\multirow{10}{*}{\textbf{Rewriting}}      & Paraphrase the input text. \\
                                         & Restate the following input text in your own words. \\
                                         & Rewrite the input text without changing its meaning. \\
                                         & Rephrase using different vocabulary. \\
                                         & Restate the input text using your own vocabulary. \\
                                         & Reconstruct the input text with new words but the same meaning. \\
                                         & Rewrite the input text while keeping its meaning. \\
                                         & Paraphrase. \\
                                         & Rewrite the text in a way that sounds different but means the same. \\
                                         & Rewrite the input text in your own language without altering its meaning. \\ \hline
\multirow{7}{*}{\textbf{Proofreading}}   & Provide a proofread version of the input text. \\
                                         & Generate a proofread version of the provided input text. \\
                                         & Offer a revised and proofread version of the given input text. \\
                                         & Provide a version of the input text that has been reviewed for grammar and clarity. \\
                                         & Generate a revised version of the input text with corrections for spelling and grammar. \\
                                         & Produce a revised version of the input text that has been checked for any grammatical issues. \\
                                         & Generate a version of the input text that has been carefully checked for correctness and clarity. \\ \hline
\multirow{10}{*}{\textbf{Translation}}    & Translate into Chinese. \\
                                         & Change the following English input text into Chinese. \\
                                         & Convert the following input text into Chinese. \\
                                         & Provide a Chinese version of the input text. \\
                                         & Translate the following text into Chinese. \\
                                         & Turn the input text into Chinese. \\
                                         & Translate into Chinese characters. \\
                                         & Rewrite this input text in Chinese. \\
                                         & Translate the text into Chinese. \\
                                         & Convert into Chinese. \\ \hline
\multirow{8}{*}{\textbf{Style Transfer}} & Rewrite the input text in a formal and polite tone. \\
                                         & Rephrase the input text in a formal and respectful manner. \\
                                         & Please rewrite the input text in a casual and conversational tone. \\
                                         & Reword the input text in a more casual and friendly tone. \\
                                         & Please rewrite the input text in an emotional and dramatic tone. \\
                                         & Transform the input text into a more emotional and dramatic tone. \\
                                         & Rephrase the input text in a clear and concise manner. \\
                                         & Please rewrite the input text in a direct and concise tone. \\ \hline \hline
\end{tabular}
}
\caption{Prompts utilized for rewriting, proofreading, translation, and style transfer tasks}
\label{Table_taskprompts}
\end{table*}

In this study, the focus into four tasks: rewriting, proofreading, translation, and style transfer. The instruction templates used for each task are provided in Table~\ref{Table_taskprompts}.

\begin{table*}[t]
\renewcommand{\arraystretch}{1.2}
\centering
\resizebox{1\textwidth}{!}{%
\begin{tabular}{clc}
\hline \hline
\textbf{Input Task}                          & \textbf{Decomposed Question}                                                                   & \textbf{Desired Answer} \\ \hline
\multirow{2}{*}{\textbf{Reasoning}}          & Is the target text a multiple-choice question?                                                  & Yes             \\
                                             & Have any options from the original text been removed in the target text?                       & No              \\ \hline
\multirow{2}{*}{\textbf{Code Generation}}    & Is the target text a code generation instruction?                                              & Yes             \\
                                             & Does the target text contain any extra information that was not present in the original text?  & No              \\ \hline
\multirow{2}{*}{\textbf{Math}}               & Is the target text a math problem?                                                             & Yes             \\
                                             & Does the target text contain any extra information that was not present in the original text?  & No              \\ \hline
\multirow{2}{*}{\textbf{Bias Detection}}     & Is the target text composed of a situation description, a question, or multiple-choice options? & Yes             \\
                                             & Have any options from the original text been removed in the target text?                       & No              \\ \hline
\multirow{2}{*}{\textbf{Question Answering}} & Is the target text composed of a passage and a question?                                       & Yes             \\
                                             & Does the target text end with a question?                                                      & Yes             \\ \hline
\textbf{+ \textit{Translation}}                       & Is the target text in LANGUAGE?                                                                & Yes             \\ \hline \hline
\end{tabular}
 }
\caption{Decomposed questions for LLM-based evaluation}
\label{table_question}
\vspace{-5mm}
\end{table*}

\section{Decomposed questions for LLM-based Evaluation}
\label{C}
As explained in Section~\ref{3.3}, we conduct LLM-based evaluation to assess how well the LLM follows instructions.
The decomposed questions for each input task can be found in Table~\ref{table_question}.
In the case of an instruction task being translation, an additional question corresponding to the translation task is included.

\clearpage

\end{document}